\pdfoutput=1
\documentclass[lettersize,journal]{IEEEtran}

\usepackage{amsmath,amsfonts,amssymb}
\usepackage{array}
\usepackage{booktabs}
\usepackage{cite}
\usepackage{graphicx}
\usepackage{textcomp}
\usepackage{stfloats}
\usepackage{flafter}
\usepackage{url}
\usepackage{makecell}
\usepackage[hidelinks]{hyperref}
\usepackage[capitalize,noabbrev]{cleveref}

\graphicspath{{figures/}}
\crefname{figure}{Fig.}{Figs.}
\Crefname{figure}{Fig.}{Figs.}
\crefname{table}{Table}{Tables}
\Crefname{table}{Table}{Tables}
\crefformat{equation}{(#2#1#3)}
\Crefformat{equation}{(#2#1#3)}

\newcommand{\Pv}{P_{\mathrm{v}}}
\newcommand{\Phat}{\hat{P}_{\mathrm{v}}}
\newcommand{\pBH}{\hat{P}_{BH}}

\begin{document}

\title{Cross-Material Support Transfer for Core-Loss Prediction Under Waveform Covariate Shift}

\author{Cong~Yao and~Chunye~Gong%
\thanks{\emph{(Corresponding author: Chunye Gong.)}}
\thanks{Cong Yao is with the School of Computer Science and Technology, Changsha University of Science and Technology, Changsha 410076, China (e-mail: ycking@stu.csust.edu.cn).}
\thanks{Chunye Gong is with the College of Computing, National University of Defense Technology, Changsha 410073, China, also with the National Supercomputer Center in Tianjin, Tianjin 300457, China, and also with the Laboratory of Digitizing Software for Frontier Equipment, National University of Defense Technology, Changsha 410073, China (e-mail: gongchunye@nudt.edu.cn).}}

\markboth{Preprint, 2026}%
{Yao and Gong: Cross-Material Support Transfer Under Waveform Covariate Shift}

\maketitle

\begin{abstract}
Power magnetic materials are characterized on the sinusoidal and triangular waveforms that excitation hardware conveniently produces, whereas deployed converters expose cores to trapezoidal, PWM-shaped flux trajectories, so loss models must predict exactly where their training data are thinnest. The final test of the MagNet Challenge embeds a deliberately extreme instance of this characterization--deployment mismatch: for material~D, trapezoids form $16.4\%$ of the test set but only $1.4\%$ of the training set. The 95th-percentile relative error, hereafter $p95$, of the best submission, built on sequential transfer learning, stalled at $15.9\%$, the worst among the five materials. This paper shows that the obstacle is missing information under covariate shift rather than class imbalance, and that the missing support can be borrowed from sibling materials instead of being extrapolated. Controlled experiments first refute the imbalance reading: four standard remedies fail, and raising the trapezoidal share to the test-set level degrades accuracy further. The proposed material-identity support transfer, MIST, then trains one 2784-parameter predictor jointly on all five challenge materials. Material identity enters through feature-wise linear modulation, or FiLM, the scarce material's true-label loss is reweighted, and material~D receives no fine-tuning, so that the bias of its trapezoid-free training set is never re-installed. MIST lowers the five-seed material-D $p95$ from $20.39\pm2.03\%$ to $12.38\pm0.92\%$ and the trapezoidal-class $p95$ from $37.4\pm8.8\%$ to $15.16\pm1.69\%$, surpassing the best submission with one-sixth of its parameters and no fine-tuning stage; removing material identity at matched capacity inflates the error by an order of magnitude. These results argue that scarce materials should be characterized jointly with their siblings.
\end{abstract}


\section{Introduction}
\IEEEPARstart{P}{ower} magnetic components are present in virtually every switching converter, and their core loss is a first-order term in the efficiency and thermal budget of a power-electronic design. The push toward higher switching frequencies and wide-bandgap devices exposes cores to increasingly non-sinusoidal, PWM-shaped flux trajectories, for which loss prediction has become a design bottleneck. Design practice still relies on the Steinmetz equation \cite{steinmetz1892} and its lineage, from the modified and improved generalized forms \cite{reinert2001_mse,venkatachalam2002_igse} to relaxation-aware and composite extensions \cite{muhlethaler2012_i2gse,guillod2023_cwh,arruti2024_cigse}. This empirical toolbox was calibrated for sinusoidal excitation and loses accuracy markedly under such arbitrary waveforms.

Data-driven core-loss modeling has matured rapidly around the MagNet project and its public datasets \cite{magnet2023,why_magnet2023,magnet_ai2023}. Neural predictors now surpass the Steinmetz family by a wide margin under arbitrary excitation, whether through physics-inspired feature fusion \cite{pi_mff_cn2024,empinn2025,xiao2026_pinn}, magnetization-mechanism priors \cite{mminn2024,huang2025_hdpi}, knowledge-aware structures \cite{deng2024_kann}, or gray-box hysteresis reconstruction, which recovers the $BH$ loop as an interpretable intermediate \cite{hardcore2025,cahrnet2026}. These models are trained and deployed per material: each material receives its own network fitted to its own measurements, and successive model families compete on features, priors, and structures within this one-material-one-model paradigm. Where the data of other materials have been used at all, they enter through \emph{sequential transfer}: a model is pretrained on legacy materials and then fine-tuned on the target material's own training set. This is the route of the transfer-learning entries of the MagNet Challenge \cite{magnet_challenge2025} and of domain-adaptation schemes that align a source material with a target material \cite{chen2025_mkmmd}. In every such pipeline the final adaptation is performed on the target material's own data, so whatever that data set lacks is never recovered. Which information one material can lend to another, and through which interface, has not been examined.

\begin{figure}[!tb]
\centering
\includegraphics[width=\columnwidth]{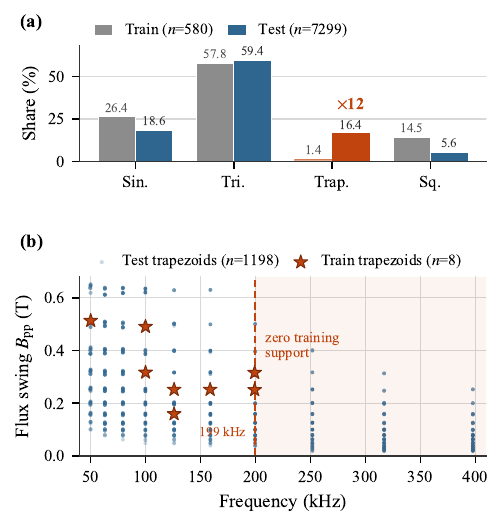}
\caption{The designed waveform challenge of material D. (a)~The trapezoidal share rises from $1.4\%$ of the training set to $16.4\%$ of the test set, a twelvefold ratio unique to this class. (b)~The eight training trapezoids (stars) stop at $199$~kHz, whereas the 1198 test trapezoids (dots) extend to $397$~kHz: the shaded high-frequency band has zero training support. This is a designed covariate shift with missing support.}
\label{fig:shift}
\end{figure}

The cost of this arrangement surfaces wherever characterization and deployment part ways. Characterization campaigns measure what the excitation hardware conveniently produces, sinusoidal and triangular waveforms, whereas deployed power converters expose the core to trapezoidal, PWM-like flux trajectories; a model fitted on the former must extrapolate to the latter. We refer to this structural gap as the \emph{characterization--deployment mismatch}: it arises from how magnetic materials are measured, so every data-driven loss model inherits it, whatever the dataset. The MagNet Challenge built its laboratory extreme into the final test and left it explicitly open \cite{magnet_challenge2025}. The \emph{waveform challenge} fell to material~D, a Fair-Rite 79 ferrite (\Cref{tab:shift}, \Cref{fig:shift}). Its trapezoids form $1.4\%$ of the training set, eight samples all within $50$--$199$~kHz, but $16.4\%$ of the test set, 1198 samples reaching $397$~kHz. The twelvefold share ratio exists for no other class and no other material, and the high-frequency trapezoidal band has no training support at all. Eight samples cannot even fill one mini-batch, so the network receives almost no trapezoid-specific supervision. Formally, this is covariate shift \cite{shimodaira2000} with \emph{missing support}: the test distribution places mass where the training distribution places almost none. Across all final submissions, black-box, gray-box, and white-box alike, spanning sixty to eleven million parameters, material~D exhibited the highest prediction error of the five materials, approaching $100\%$ in the worst case, and the highest cross-team variation. The best reported 95th-percentile ($p95$) relative error, $15.9\%$, required a 16449-parameter model with a dedicated transfer-learning stage and was still the worst among the five materials.

\begin{table}[!tb]
\caption{Waveform Composition of the Material-D Training and Test Sets}
\label{tab:shift}
\centering
\small
\renewcommand{\arraystretch}{1.2}
\begin{tabular}{l c c}
\toprule
Waveform class & Training ($n=580$) & Test ($n=7299$) \\
\midrule
Sinusoidal  & 153 (26.4\%) & 1357 (18.6\%) \\
Triangular  & 335 (57.8\%) & 4335 (59.4\%) \\
Trapezoidal & \textbf{8 (1.4\%)} & \textbf{1198 (16.4\%)} \\
Square-like & 84 (14.5\%)  & 409 (5.6\%) \\
\bottomrule
\end{tabular}
\end{table}

\begin{figure}[!tb]
\centering
\includegraphics[width=\columnwidth]{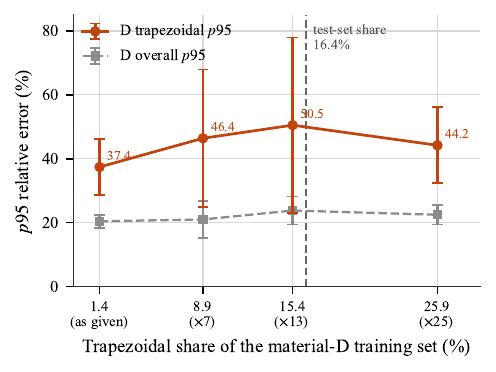}
\caption{The oversampling paradox that motivates this paper. Replicating the eight training trapezoids of material D until their share reaches or exceeds the test-set share (dashed line) makes the trapezoidal-class $p95$ worse and inflates its cross-seed variance (five-seed mean$\pm$std). If the difficulty were class imbalance, share matching would be the textbook remedy; its failure identifies the difficulty as missing information.}
\label{fig:paradox}
\end{figure}

One question decides how the waveform challenge must be treated. \emph{Is the scarce class merely under-represented, or is the information it requires absent from the training data altogether?} Under \emph{frequency imbalance}, oversampling \cite{smote2002}, reweighting, synthetic augmentation \cite{mixup2018}, or pseudo-labeling should help, as they do for imbalanced regression targets in general \cite{yang2021_dir}. Under \emph{missing information}, every method that only reshuffles, re-parameterizes, or re-weights the same 580 samples must fail, because no manipulation of a dataset can create information it does not contain. Unexpectedly for a class that is present in the training set, the difficulty turns out \emph{not} to be class imbalance. Under a controlled five-seed protocol, four single-material remedies all fail: continuous waveform descriptors in place of one-hot classes, synthetic trapezoids labeled by an improved generalized Steinmetz equation, direct oversampling of the eight trapezoids, and exact polarity-symmetry augmentation. Raising the trapezoidal share to the test-set level even makes the trapezoidal error worse (\Cref{fig:paradox}), as verified in \Cref{sec:failures}. The failure is thus a missing-information problem under covariate shift, and the imbalance toolbox does not apply to it.

The missing geometry exists by construction in material~D's siblings. The training sets of the four sibling materials, characterized on the same excitation hardware, contain 2447 \emph{measured} trapezoids, and their flux-swing/frequency footprint covers $99.3\%$ of the material-D test trapezoids within a $\pm10\%$ neighborhood. In this paper, we address the missing-information problem by introducing \emph{material-identity support transfer} (MIST). We hypothesize that \emph{the missing support can be borrowed across materials, provided the borrowing interface shares what is physically common and separates what is not}. The chain from flux density to magnetic field to hysteresis loop to loss is common to all ferrites, whereas the loop shape is material-specific and shifts the loss by orders of magnitude for the same excitation. MIST therefore trains a single shared gray-box reconstruction backbone on the joint five-material pool and injects material identity through feature-wise linear modulation (FiLM) \cite{film2018}. Per-material scale-and-shift parameters then separate the material response, while every material's trapezoids supervise the same shared representation (\Cref{sec:arch}). Two more decisions define MIST. The scarce material's loss term is reweighted, an importance correction on \emph{true} labels in the spirit of covariate-shift weighting \cite{shimodaira2000,kliep2007} (\Cref{sec:objective}). Unlike sequential transfer, MIST never fine-tunes material~D: its fine-tuning set is the trapezoid-free set whose bias the joint training escaped, and any adaptation on it brings that bias back through catastrophic forgetting \cite{mccloskey1989} (\Cref{sec:training-decisions}).

The contributions are summarized as follows.
\begin{itemize}
\item \emph{An experimentally settled diagnosis of the waveform challenge.} Four controlled single-material remedies fail and share matching degrades accuracy, which identifies the difficulty as missing information under covariate shift.
\item \emph{Material-identity support transfer (MIST).} A single 2784-parameter predictor trained on the joint A--E pool replaces five per-material models; a capacity-matched ablation shows that the gain comes from identity information, not from capacity or data volume.
\item \emph{The best reported accuracy on the open problem.} MIST attains a material-D $p95$ of $12.38\pm0.92\%$ with all five seeds below $14\%$, against $15.9\%$ for the best submission, cuts the trapezoidal-class $p95$ from $37.4\pm8.8\%$ to $15.16\pm1.69\%$, and improves the five-material aggregate from $7.89\%$ to $7.32\%$.
\item \emph{Open resources for the power electronics community.} We make the MIST algorithm code openly available to contribute to the power electronics open-source community and provide a reference for future research.
\end{itemize}

\section{Methodology}
\label{sec:method}

\subsection{Method Overview}
\label{sec:premise}
\begin{figure*}[!t]
\centering
\includegraphics[width=0.92\textwidth]{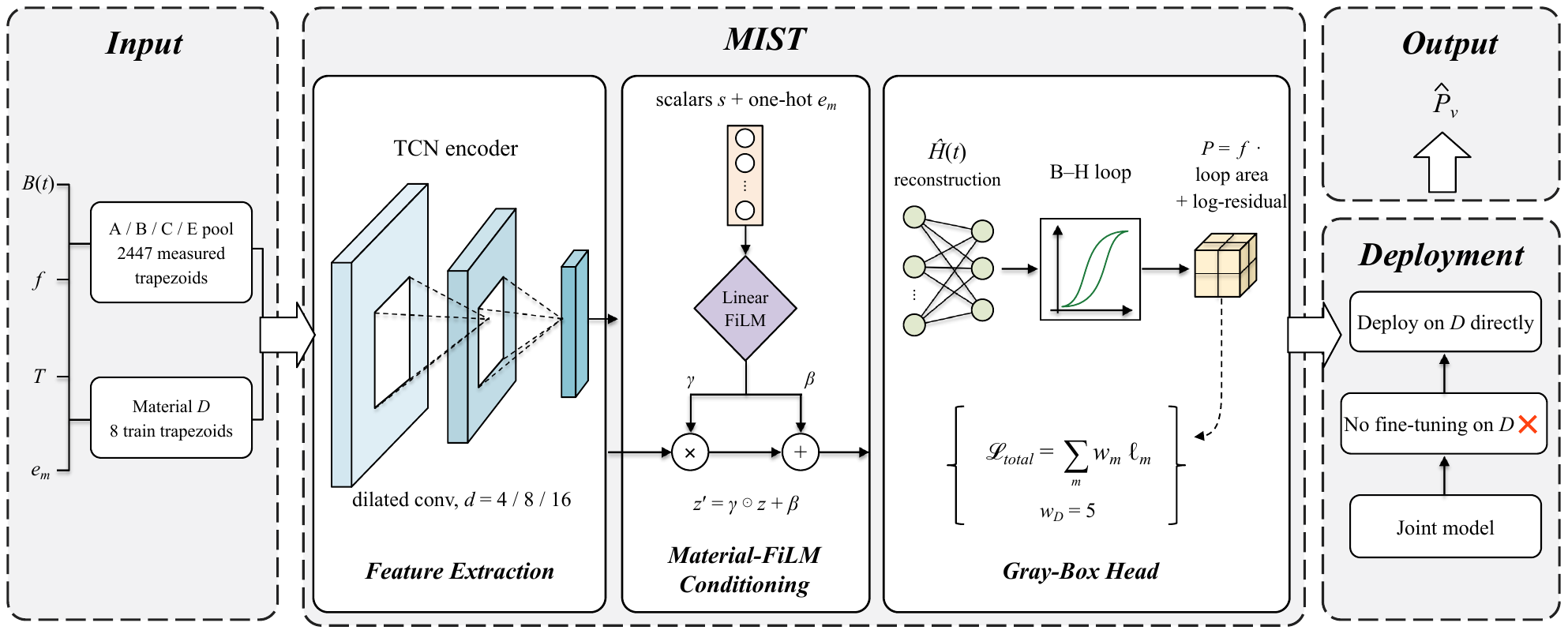}
\caption{The proposed MIST architecture. All five materials train a single 2784-parameter gray-box predictor that follows the shared chain $\mathbf{B}\rightarrow\hat{\mathbf{H}}\rightarrow BH\rightarrow\Phat$; material identity enters through the scalar$\rightarrow$FiLM conditioning pathway, the scarce material's true-label loss is reweighted ($w_\mathrm{D}=5$), and material~D is deployed directly from the joint model without fine-tuning.}
\label{fig:arch}
\end{figure*}

The architecture follows the split in the physics (\Cref{fig:arch}): the chain from flux trajectory $\mathbf{B}$ to field response $\mathbf{H}$ to hysteresis loop to loss is shared by all materials, whereas the loop shape is material-specific. A single loss predictor receives the flux waveform and follows the gray-box chain $\mathbf{B}\rightarrow\hat{\mathbf{H}}\rightarrow BH\rightarrow\Phat$ for every material. A shared reconstruction backbone thus learns the excitation-to-loop mapping from all materials at once, while a lightweight conditioning pathway separates the material response. True-label reweighting amplifies the scarce material's measured supervision, and material~D is deployed directly from the joint model without fine-tuning.

\begin{figure}[!tb]
\centering
\includegraphics[width=\columnwidth]{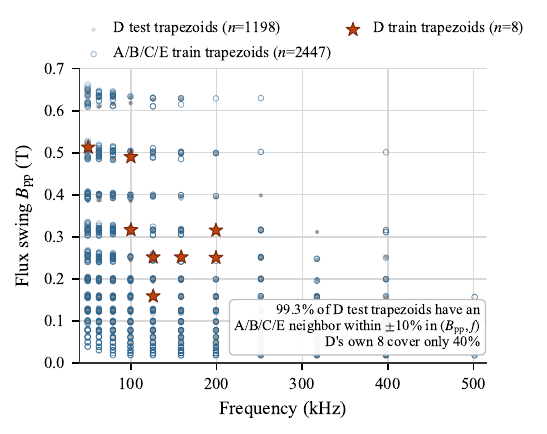}
\caption{Measured trapezoidal support in the $(f, B_{\mathrm{pp}})$ plane. The 2447 A/B/C/E training trapezoids (open circles) blanket the material-D test trapezoids (dots): $99.3\%$ of the latter have a sibling neighbor within $\pm10\%$ in both flux swing and frequency, whereas material~D's own eight trapezoids (stars) cover only $40\%$.}
\label{fig:premise}
\end{figure}

Borrowing requires something to borrow, so this premise is checked first. The training sets of materials A, B, C, and E contain 336, 952, 788, and 371 measured trapezoids respectively, 2447 in total and a $13$--$18\%$ share within each material. Their flux-swing/frequency footprints individually cover $91$--$97\%$ of the material-D test-trapezoid box. Jointly, $99.3\%$ of the material-D test trapezoids have a neighbor within $\pm10\%$ in both flux swing and frequency among the sibling materials' training trapezoids, whereas material~D's own eight trapezoids cover only $40\%$ (\Cref{fig:premise}). The cross-material route transfers \emph{measured} loop geometry, the one ingredient that no manipulation of material~D's own samples can supply (\Cref{sec:failures}). The zero-shot experiment of \Cref{sec:generalization} tests whether this borrowed support alone produces the gain, and the band decomposition of \Cref{sec:band} checks whether the gain lands where the support is borrowed.

\subsection{Shared Gray-Box Backbone With Material-Identity FiLM Conditioning}
\label{sec:arch}
The backbone is the CAHR-Net predictor \cite{cahrnet2026}, which follows the interpretable chain $\mathbf{B}\rightarrow\hat{\mathbf{H}}\rightarrow BH\rightarrow\Phat$ of the gray-box family \cite{hardcore2025}. Its encoder is a three-layer dilated temporal-convolutional network \cite{tcn2018} with kernel width nine, dilations 4/8/16, and circular padding. It maps five time-series channels, namely the normalized flux waveform, its first and second differences, and two shape features, to the reconstructed field $\hat{\mathbf{H}}$. The loop-area head integrates $\hat{\mathbf{H}}$ against the input flux, and a two-layer residual head refines the result in the logarithmic loss domain. A scalar vector $\mathbf{s}$ conditions the encoder through a linear FiLM branch; its eleven entries are frequency, temperature, four one-hot waveform indicators, flux swing, mean slew rate, their logarithms, and the sample time. MIST appends material identity as a five-dimensional one-hot vector $\mathbf{e}_m$ to $\mathbf{s}$ and routes it through this existing pathway, whose FiLM block \cite{film2018} modulates the intermediate representation $\mathbf{z}$ feature-wise,
\begin{equation}
\label{eq:film}
\mathbf{z}' = \boldsymbol{\gamma}(\mathbf{s},\mathbf{e}_m)\odot \mathbf{z} + \boldsymbol{\beta}(\mathbf{s},\mathbf{e}_m),
\end{equation}
so the linear scalar branch learns per-material scale-and-shift pairs, a \emph{material FiLM}, without any change to the backbone topology. Two bookkeeping decisions keep the mixed pool physically consistent. Per-material input/output normalization is retained from the single-material baseline. The per-sample physical bounds $b_{\lim}$ and $h_{\lim}$, the flux and field scales used to normalize each sample, are passed into the loop-area head, so that within a mixed-material batch each sample's loss reconstruction
\begin{equation}
\label{eq:physhead}
\pBH = f\oint \hat{H}\,\mathrm{d}B \approx f\, h_{\lim} b_{\mathrm{pp}} \sum_{t} \hat{H}_t\,\Delta B_t
\end{equation}
is evaluated on its own physical scale, where $b_{\mathrm{pp}}$ is the peak-to-peak flux swing and $\Delta B_t$ the normalized flux increment. The complete model has 16 scalar inputs and \textbf{2784} parameters, one network replacing the five per-material baselines and their $5\times1874=9370$ parameters, a $70\%$ reduction. Because the encoder width of this topology is tied to the scalar count, the five identity columns add 910 parameters; the capacity-matched control of \Cref{sec:ablation} reproduces this width without identity information.

\subsection{True-Label Reweighted Objective}
\label{sec:objective}
Material~D contributes only $3.3\%$ of the joint pool, so its gradient signal is diluted where accuracy matters most. The correction is an importance weight on the \emph{true-label} loss terms of the scarce material,
\begin{equation}
\label{eq:objective}
\mathcal{L} = \sum_{m\in\{\mathrm{A},\ldots,\mathrm{E}\}} w_m \, \mathbb{E}_{(\mathbf{B},\mathbf{s},\Pv)\sim\mathcal{D}_m}\!\left[\ell\!\left(\ln\Phat,\ln\Pv\right)\right],
\end{equation}
with $w_\mathrm{D}=K$ and $w_m=1$ otherwise, the simplest instance of covariate-shift importance weighting \cite{shimodaira2000,kliep2007}. Here $\ell$ is the staged objective of the backbone \cite{cahrnet2026}, which combines the log-loss error with a field-reconstruction error whose weight decays over training, so that $\hat{\mathbf{H}}$ is learned first and the loss target dominates at convergence. The distinction from the pseudo-label route of \Cref{sec:failures} is categorical: reweighting amplifies measured labels, whereas pseudo-labeling invents them. A sensitivity sweep (\Cref{sec:dose}) locates a shallow optimum at $K\in[3,5]$; we adopt $K=5$.

\subsection{Training and Deployment Procedure}
\label{sec:training-decisions}
Two training decisions remain, and both are grounded in the experiments of \Cref{sec:results}. First, the joint model is trained long, 10000 epochs at batch size 512 under AdamW \cite{adamw2019} with cosine scheduling \cite{sgdr2017}, because accuracy is still improving on every axis at the 3000-epoch operating point (\Cref{sec:dose}). Second, \emph{material~D is never fine-tuned}. Its adaptation set is the trapezoid-free 580-sample set whose bias the joint training escaped. Any per-material adaptation on it, full or head-only, at any strength, pulls the conditioning parameters back toward the biased distribution and erases the borrowed trapezoidal knowledge, an instance of catastrophic forgetting \cite{mccloskey1989}. The fine-tuning strength sets how much of the borrowed knowledge is forgotten (\Cref{sec:ft}), and the correct strength for the scarce material is zero: it is deployed directly from the joint model.

\section{Experimental Setup}

\subsection{Dataset, Metrics, and Evaluation Protocol}
The MagNet Challenge final protocol provides a training set and a sealed test set for each of the five materials A--E. Each training sample is a measured tuple: a single-period flux-density waveform $\mathbf{B}=\{B_t\}_{t=1}^{N}$ with $N=1024$, an excitation frequency $f$, a temperature $T$, and the volumetric loss $\Pv$. All experiments use the official training/test division; the material-D composition is given in \Cref{tab:shift}. For each sample the relative error $e=|\Phat-\Pv|/\Pv$ is computed \cite{tofallis2015}, and three statistics are reported: the per-material $p95$, the material-D trapezoidal-class $p95$ over the 1198 test trapezoids, and the five-material aggregate $p95$, the mean of the per-material values. Unless stated otherwise, every number is a mean$\pm$std over five independent seeds.

All experiments follow three evaluation rules, fixed before any method was tried. \emph{(i) Three targets must improve together}: the material-D $p95$, the material-D trapezoidal-class $p95$, and the five-material aggregate $p95$; a gain on D that is paid for by the aggregate is not accepted. The trapezoidal-class $p95$ is the target metric, because the aggregate is dominated by the $59\%$ triangular share and can mask the mechanism under study. \emph{(ii) Headline numbers are five-seed mean$\pm$std}; single-seed values are reported only as supplementary evidence. \emph{(iii) Every auxiliary loss term carries a zero-weight control}: the term is attached with zero weight, so that pipeline confounds are separated from the effect of the term itself.

\subsection{Baseline and Compared Methods}
\label{sec:compared}
The baseline is the strongest available single-material configuration: five independent per-material CAHR-Net models \cite{cahrnet2026}, each with 1874 parameters. Each model builds on the gray-box reconstruction chain of HARDCORE \cite{hardcore2025} with a temporal-convolutional encoder \cite{tcn2018} and is trained with AdamW \cite{adamw2019} under cosine scheduling \cite{sgdr2017} at batch size 512. Over five seeds it attains an aggregate $p95$ of $7.89\pm0.36\%$, a material-D $p95$ of $20.39\pm2.03\%$, and a material-D trapezoidal $p95$ of $37.4\pm8.8\%$.

The compared methods fall into three data regimes, which the tables mark explicitly. The \emph{single-material regime} trains one model per material on that material's data alone. It comprises the baseline and the four single-material remedies of the diagnostic study, each of which manipulates only material~D's own 580 training samples (\Cref{sec:failures}). It also includes the first-ranked submission HARDCORE \cite{hardcore2025}, the ancestor of the baseline, which we re-train with its official code and data under the five-seed protocol. The \emph{sequential-transfer regime} pretrains on the ten legacy MagNet materials and then fine-tunes on each final material; the two strongest such submissions are quoted from the official report \cite{magnet_challenge2025}, since their released artifacts do not permit a controlled re-training: MagLearn from Bristol \cite{maglearn2024}, whose material-D result used a 16449-parameter model and whose pretrained base models for materials D and E are not released, and Fuzhou, whose training code is not released. The \emph{joint regime} trains the single MIST model of \Cref{sec:method} on the A--E pool in three configurations: MIST w/o RW without reweighting; MIST with reweighting at $w_\mathrm{D}=5$, the proposed configuration; and MIST+FT, which follows joint pretraining by per-material fine-tuning on A/B/C/E with none on D. Two ablations complete the set on identical data: \emph{pooling} removes the five identity columns and the parameters they feed, returning to the 1874-parameter single-material topology. The \emph{dummy} ablation retains the columns but fills them with a constant uniform vector, matching the 2784-parameter capacity of MIST exactly.

\subsection{Implementation Details and Result Integrity}
The joint model is trained with AdamW \cite{adamw2019} under cosine scheduling \cite{sgdr2017} at batch size 512 for 10000 epochs, with the reweighted objective \cref{eq:objective}. Three checks protect the integrity of the results. First, a mathematically equivalent second implementation, differing only in scalar column order and RNG stream, reproduces the method but shifts the same-seed material-D $p95$ across a $3.7$-percentage-point range, from $12.20$ to $15.91\%$ at seed~0. A nominal seed is therefore not a reproducible draw, which motivates the five-seed mean$\pm$std reporting rule. Second, the test directories are isolated from training and the metric code is shared verbatim with the baseline evaluation, so test labels never enter the training pipeline. Third, all zero-weight controls of the diagnostic study (\Cref{sec:failures}) are retained in the joint experiments.

\section{Experimental Results and Discussion}
\label{sec:results}

\subsection{Diagnosis: Single-Material Remedies Under Zero-Weight Control}
\label{sec:failures}
None of the four single-material remedies improves the trapezoidal axis, and matching the test-set share makes it worse; this section establishes both results under controlled conditions. A batch-size sweep of the baseline already localizes the failure. Reducing the batch size from 1024 to 512 to 256 monotonically improves materials A, B, C, and E, while every degradation concentrates in material~D, $17.96 \rightarrow 18.57 \rightarrow 20.63\%$ at seed~0, and the five-seed means show the same trend. The worst material-D samples are exclusively trapezoidal, with a trapezoidal $p95$ between $37$ and $60\%$ and a cross-seed standard deviation up to $31$ percentage points, whereas the sinusoidal and triangular classes remain stable. The material-D failure is thus seed-driven, class-localized, and robust to the training regime. A hard loss landscape would hurt every class and respond to the batch size; what is observed instead hurts one class and follows the seed, which is the mark of absent supervision.

The four routes of \Cref{sec:compared} share one premise: no external real data are introduced; each manipulates only material~D's own 580 training samples. \Cref{tab:donly} summarizes the outcomes on the two material-D axes.

\begin{table*}[!tbp]
\caption{Single-Material Routes on the Material-D Waveform Challenge (Five Seeds Unless Noted; Baseline in Each Block for Reference)}
\label{tab:donly}
\centering
\small
\setlength{\tabcolsep}{5pt}
\renewcommand{\arraystretch}{1.2}
\begin{tabular}{p{5.0cm} c c p{6.6cm}}
\toprule
Route & D $p95$ (\%) & D trap.\ $p95$ (\%) & Outcome \\
\midrule
Baseline (per-material model) & $20.39\pm2.03$ & $37.4\pm8.8$ & reference \\
\midrule
\multicolumn{4}{@{}l}{\textit{(a) Representation replacement (three seeds, bs1024; one-hot at bs1024: $17.96$, seed~0)}}\\
Continuous waveform descriptors & $28.18\pm1.26$ & $42.9$--$63.5$ & variance halved, mean collapses \\
\midrule
\multicolumn{4}{@{}l}{\textit{(b) Synthetic trapezoids with iGSE pseudo-labels}}\\
Zero-weight control ($\lambda=0$) & $19.88\pm2.32$ & --- & reproduces baseline \\
Pseudo-labels ($\lambda=0.05$) & $26.77\pm2.72$ & --- & degradation attributable to pull \\
Pseudo-labels ($\lambda=0.25$) & $24.77\pm2.11$ & $39.5\pm3.8$ & trap.\ variance drops, all other classes degrade \\
\midrule
\multicolumn{4}{@{}l}{\textit{(c) Oversampling the eight trapezoids by replication (\Cref{fig:paradox})}}\\
$\times 7$ (share $8.9\%$) & $20.97\pm5.85$ & $46.4\pm21.5$ & worse, unstable \\
$\times 13$ (share $15.4\%$, matches test) & $23.77\pm4.34$ & $50.5\pm27.5$ & \textbf{refutes class imbalance} \\
$\times 25$ (share $25.9\%$) & $22.44\pm3.02$ & $44.2\pm11.9$ & worse \\
\midrule
\multicolumn{4}{@{}l}{\textit{(d) Exact symmetry augmentation}}\\
Polarity flip (all materials) & $19.00\pm4.45$ & $49.05$ (mean) & insignificant mean shift, variance $\times 2.7$ \\
\bottomrule
\end{tabular}
\par\vspace{2pt}
{\footnotesize\raggedleft Configurations: the baseline is trained with batch size 512. Block (a) replaces the one-hot waveform classes by four continuous descriptors at the same 1874 parameters, uses batch size 1024 and three seeds, and reaches an aggregate $p95$ of $10.07\pm0.23\%$. In block (b) the zero-weight control attaches the synthetic branch at zero weight, and the pseudo-label rows add 1194 synthetic trapezoids with loss-domain pseudo-labels at weight $\lambda$; the zero-weight control reproduces the baseline seed by seed, so the degradation is caused by the pseudo-label pull itself, not by the pipeline; the trapezoidal-class statistic was recorded for the strongest setting only. In block (d) the circular-shift symmetry is already exact in the backbone, so only the polarity flip is added; the paired five-seed differences against the baseline are statistically insignificant, and the per-seed trapezoidal $p95$ spans $17.9$--$127.9\%$. ``trap.'' denotes the trapezoidal class.\par}
\end{table*}

\emph{Representation replacement.} Replacing the four one-hot waveform classes by four continuous descriptors, namely waveform form factor, crest factor, duty, and fundamental ratio, at identical parameter count halves the cross-seed variance but raises the material-D $p95$ to $28.18\pm1.26\%$ and roughly doubles the trapezoidal-class error. Under one-hot encoding a test trapezoid at least inherits the coarse bias of the square-like class, whereas under continuous coordinates it receives a precise descriptor position at which the training set has no support. Interpolation in descriptor space is not interpolation of the underlying physics: re-parameterizing the same 580 samples cannot restore missing support.

\emph{Physics-based pseudo-labels.} A per-temperature improved-generalized-Steinmetz (iGSE) fit \cite{venkatachalam2002_igse} of the material-D training set ($\alpha\in[1.45,1.90]$, $\beta\in[2.75,2.95]$) generates 1194 synthetic trapezoidal waveforms with weak loss labels, attached to the loss pathway with weight $\lambda$. The zero-weight control ($\lambda=0$) reproduces the baseline seed by seed, so the observed degradation at $\lambda>0$ is caused by the pseudo-label pull itself. The prior anchors the extrapolation region and cuts the trapezoidal variance by two-thirds, but the $15$--$25\%$ systematic bias of iGSE propagates through the shared weights into the regions that have measured supervision. The aggregate deteriorates: the bias that the invented labels import outweighs the variance they remove.

\emph{Oversampling.} Replicating the eight trapezoids by factors of 7, 13, and 25 raises the trapezoidal share to $8.9\%$, to $15.4\%$, which matches the test set, and to $25.9\%$. Every setting degrades the trapezoidal-class $p95$ itself, from $37.4\pm8.8\%$ to $44$--$51\%$, while inflating its variance up to $\pm 28$ percentage points (\Cref{fig:paradox}). This negative result settles the diagnosis: if the difficulty were class imbalance, matching the test-set share would be the textbook remedy \cite{smote2002}, yet it makes the trapezoidal error strictly worse. Eight samples repeated thirteen times still describe eight low-frequency loop geometries, and replication only concentrates optimization pressure on them.

\emph{Exact symmetry augmentation.} Circular time shift is already exact in the backbone through its circular padding, so only polarity reversal of the flux waveform is tested, applied to all materials. It leaves the five-seed means almost unchanged, aggregate $7.65\pm0.88$ against $7.89\%$ and material-D $19.00\pm4.45$ against $20.39\%$. It multiplies the cross-seed variance by $2.7$, however, with the per-seed material-D trapezoidal $p95$ ranging from $17.9$ to $127.9\%$. The flip creates no new loop geometry: the eight trapezoids become sixteen near-duplicates, so a symmetry that the data already obey cannot inject missing support and only redistributes the seed-to-seed variance.

The four routes span the standard imbalance toolbox and fail in a mutually consistent pattern; most tellingly, share matching, the one manipulation that directly tests the imbalance hypothesis, refutes it outright. The diagnosis is that the material-D failure is \emph{missing information under covariate shift}, and the missing high-frequency trapezoidal loop geometry can only enter the model as \emph{real measured data from outside material~D}.

\subsection{Main Comparison}
\label{sec:main}
\begin{figure}[!tb]
\centering
\includegraphics[width=\columnwidth]{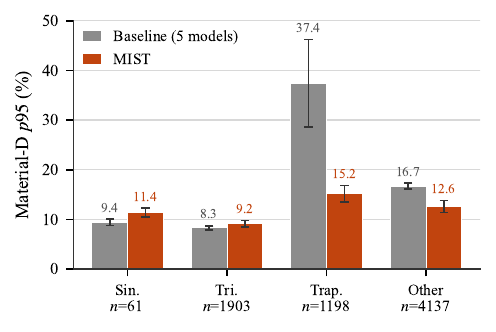}
\caption{Material-D $p95$ by waveform class (five-seed mean$\pm$std). MIST cuts the trapezoidal class from $37.4\%$ to $15.2\%$ and the mixed \emph{other} class from $16.7\%$ to $12.6\%$, while the well-supported sinusoidal and triangular classes concede at most two percentage points.}
\label{fig:main}
\end{figure}

MIST improves all three targets at once, and by the widest margin on the trapezoidal axis. \Cref{tab:main} consolidates the comparison across the three data regimes, and \Cref{fig:main} shows the per-class decomposition. The proposed configuration, MIST with $2784$ parameters and no fine-tuning, attains a material-D $p95$ of $12.38\pm0.92\%$ with every one of the five seeds below $14\%$ ($12.52/11.86/12.79/11.17/13.58$), a trapezoidal-class $p95$ of $15.16\pm1.69\%$, and an aggregate of $7.32\pm0.19\%$. Against the single-material baseline, material~D falls by $8.0$ percentage points. More importantly, the trapezoidal target, the axis that every single-material route of \Cref{sec:failures} failed to move, falls to $40\%$ of its baseline value with its variance reduced fivefold. One network at $30\%$ of the total parameter budget replaces five and still improves the aggregate.

Against the challenge submissions, MIST surpasses MagLearn's $15.9\%$ material-D result, the best value reported on the axis the organizers identified as the open problem \cite{magnet_challenge2025}, by $3.5$ percentage points with about one-sixth of its parameters. MagLearn obtained that value by sequential transfer, pretraining on the legacy materials and fine-tuning on material~D. MIST consumes comparable external data but never adapts on material~D's own set, and \Cref{sec:ft} shows that this omission is what preserves the borrowed geometry. HARDCORE, re-trained under the same five-seed protocol, reaches $23.44\pm3.79\%$ on material~D. On the aggregate, MIST's $7.32\%$ is lower than the $7.78\%$ and $7.94\%$ of MagLearn and Fuzhou, computed from their reported per-material values, and than the $8.37\pm0.87\%$ of the re-trained HARDCORE ($7.84\%$ self-reported). It does so with a single 2784-parameter network in place of five models of 1755 to 90653 parameters each.

The shared representation redistributes accuracy toward the scarce material. Each well-supported material concedes a mild amount (A $5.68\rightarrow6.65$, B $2.23\rightarrow3.21$, C $3.82\rightarrow4.81$, E $7.33\rightarrow9.55$), the open-problem axis improves by far more, and the aggregate improves. The concession is partly recoverable: fine-tuning the well-supported materials wins some of it back at a cost on material~D that \Cref{sec:ft} quantifies. \Cref{fig:main} decomposes the material-D error by waveform class: the gain concentrates on the trapezoidal and mixed classes, while the sinusoidal and triangular classes concede at most two percentage points.

\begin{table*}[!tbp]
\caption{Main Comparison on the MagNet Final A--E Protocol: Material-D, Aggregate, and Material-D Trapezoidal-Class $p95$ (\%) (Five-Seed Mean$\pm$Std Unless Noted)}
\label{tab:main}
\centering
\small
\setlength{\tabcolsep}{5pt}
\renewcommand{\arraystretch}{1.2}
\begin{tabular}{p{7.4cm} c c c c}
\toprule
Method & Params & D $p95$ & Agg.\ $p95$ & D trap.\ $p95$ \\
\midrule
\multicolumn{5}{@{}l}{\textit{(a) Single-material regime (each model sees only its own material)}}\\
Baseline: five per-material CAHR-Net models \cite{cahrnet2026} & $5\times1874$ & $20.39\pm2.03$ & $7.89\pm0.36$ & $37.4\pm8.8$ \\
HARDCORE \cite{hardcore2025}$^{\P}$ & $5\times1755$ & $23.44\pm3.79$ & $8.37\pm0.87$ & $40.8\pm6.9$ \\
\midrule
\multicolumn{5}{@{}l}{\textit{(b) Sequential-transfer regime (pretraining on the legacy materials, then fine-tuning on each material including D)}}\\
MagLearn \cite{maglearn2024}$^{\ddag}$ & 16449$^{\ast}$ & 15.90 & 7.78 & 22.9 \\
Fuzhou$^{\ddag}$ & $5\times8914$ & 20.70 & 7.94 & 27.6 \\
\midrule
\multicolumn{5}{@{}l}{\textit{(c) Joint regime (this work, one shared model trained on the A--E pool, no fine-tuning on D)}}\\
MIST w/o RW & 2784 & $14.92\pm1.41$ & $7.64\pm0.35$ & $20.98\pm5.79$ \\
\textbf{MIST (proposed)} & \textbf{2784} & $\mathbf{12.38\pm0.92}$ & $\mathbf{7.32\pm0.19}$ & $\mathbf{15.16\pm1.69}$ \\
MIST+FT$^{\|}$ & 2784 & $14.01\pm1.93$ & $7.24\pm0.50$ & $18.08\pm4.18$ \\
\midrule
\multicolumn{5}{@{}l}{\textit{(d) Ablations of identity conditioning (same A--E pool, 3000 epochs; row 1 keeps the identity, dummy is capacity-matched)}}\\
MIST w/o RW, 3000 epochs & 2784 & $16.61\pm1.58$ & $8.84\pm0.35$ & $21.3\pm3.1$ \\
Pooling$^{\S}$ & 1874 & $145.5\pm16.3$ & $61.51\pm5.92$ & $140\pm24$ \\
Dummy$^{\dag}$ & 2784 & $125.12\pm33.62$ & $54.65\pm8.65$ & $131.8\pm32.3$ \\
\bottomrule
\end{tabular}
\par\vspace{2pt}
{\footnotesize\raggedleft Agg.\ is the mean of the five per-material $p95$ values; the per-material values of this work are given in \Cref{sec:main}. $^{\P}$Official 1st; re-trained by us with the official code and data, as released, under our five-seed protocol; the self-reported single run is 22.2 on material D with aggregate 7.84 \cite{magnet_challenge2025}. $^{\ddag}$Quoted from \cite{magnet_challenge2025} (MagLearn: official 3rd and best material-D value; Fuzhou: official 2nd): single runs under the authors' own implementations, external references rather than a controlled ranking; the trapezoidal-class value is computed by us from the predictions the team submitted to the challenge, with the waveform classifier of this paper. $^{\ast}$MagLearn used 90653-parameter models for materials A--C and a 16449-parameter transfer-learned model for D--E; its material-D value (15.90) stems from the latter. $^{\|}$Per-material fine-tuning of A/B/C/E after joint pretraining, D untouched. $^{\S}$No identity columns; removing them also removes the parameters they feed, because the encoder width of this topology is tied to the scalar count (\Cref{sec:arch}). $^{\dag}$Identity columns present but filled with a constant, i.e., zero information, at the full 2784-parameter width; three seeds. Blocks (a)--(d) mark the data regime: single-material rows use only each material's own data, sequential-transfer rows pretrain on the ten legacy materials and fine-tune on D, and joint rows train one model on the A--E pool without fine-tuning on D.\par}
\end{table*}

\subsection{Ablation Study}
\label{sec:ablation}
Identity information, not the 910 additional parameters, carries the gain. Block~(d) of \Cref{tab:main} compares three configurations on identical data and an identical 3000-epoch schedule. The \emph{pooling} ablation removes the five identity columns and returns the network to the 1874-parameter single-material topology. It inflates the material-D $p95$ from $16.61\pm1.58\%$ to $145.5\pm16.3\%$, roughly an order of magnitude. The same flux trajectory produces losses that differ by orders of magnitude across materials, so an unconditioned shared network can only regress toward a cross-material mean. The \emph{dummy} ablation restores the five columns but fills them with a constant uniform vector. It restores the full 2784-parameter width of MIST yet recovers almost nothing, $125.12\pm33.62\%$, which rules out the 910 additional parameters as the source of the gain. Only the informative identity restores accuracy. The gain of cross-material support transfer therefore comes from material-identity \emph{information} delivered through the FiLM pathway: the dummy row holds capacity and input dimensionality fixed, and all three rows share the same data. This also extends the single-material finding of \cite{cahrnet2026}, where condition modulation yielded no measurable benefit \emph{within} one material: the mechanism matters once knowledge must be shared across materials.

\subsection{Error Decomposition by Frequency Band}
\label{sec:band}
\begin{figure}[!tb]
\centering
\includegraphics[width=\columnwidth]{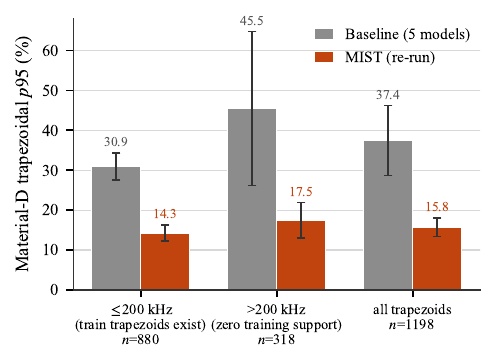}
\caption{Material-D trapezoidal $p95$ by frequency band (five-seed mean$\pm$std). The baseline degrades and destabilizes above $200$~kHz, where material~D has no training trapezoid; MIST closes most of that gap.}
\label{fig:band}
\end{figure}

If the gain comes from borrowed support, it should land where material~D's own support is missing; generic regularization would spread it evenly. \Cref{fig:band} splits the 1198 material-D test trapezoids at the upper edge of the training trapezoids' frequency range (\Cref{fig:shift}(b)): 880 trapezoids up to $200$~kHz, where material~D's own eight training trapezoids reside, and 318 above $200$~kHz, where the training support is empty. The per-sample errors of MIST come from a five-seed re-run of the proposed configuration with per-sample logging, which reproduces \Cref{tab:main} within one standard deviation (material-D $p95$ $12.67\pm0.61\%$). The baseline degrades sharply where support is absent, from $30.9\pm3.4\%$ below $200$~kHz to $45.5\pm19.3\%$ above, with the cross-seed spread widening nearly sixfold, as expected when a model extrapolates. MIST removes most of this gap, $14.3\pm2.1\%$ versus $17.5\pm4.4\%$: the zero-support band improves by $28$ percentage points and its seed spread collapses, whereas the supported band improves by $17$. The largest and most stable gain occurs where material~D's own data cannot help and the sibling materials' measured trapezoids do exist (\Cref{fig:premise}). The gain therefore tracks the borrowed support coverage; a smoother fit of material~D's own samples would spread it evenly.

\subsection{Sensitivity to the Reweighting Factor and Training Length}
\label{sec:dose}
\begin{figure}[!tb]
\centering
\includegraphics[width=\columnwidth]{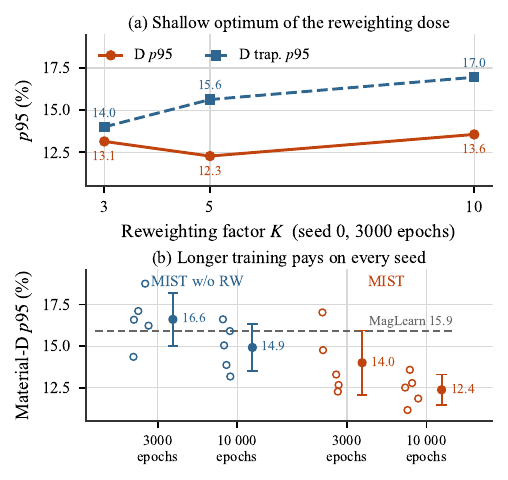}
\caption{(a)~A shallow single-valley response curve over the reweighting factor $K$ (seed~0, 3000 epochs) indicates a genuine mechanism rather than a tuned constant. (b)~Longer training improves both configurations on every seed (open circles: individual seeds; filled: mean$\pm$std); already at 3000 epochs the reweighted five-seed mean reaches the level of MagLearn's $15.9\%$.}
\label{fig:dose}
\end{figure}

The reweighting factor has a shallow, smooth optimum (\Cref{fig:dose}). At seed~0 the material-D $p95$ moves through $13.14$ ($K=3$), $12.28$ ($K=5$), and $13.56$ ($K=10$), with the trapezoidal axis at $14.0/15.6/17.0\%$, and $K=10$ begins to penalize materials A and E ($8.70$ and $12.13\%$). A smooth single-valley response curve is what a genuine mechanism produces; the pseudo-label route in \Cref{tab:donly}(b) degrades non-monotonically instead. Training length matters as well: extending from 3000 to 10000 epochs improves MIST w/o RW from $16.61\pm1.58$ to $14.92\pm1.41\%$ and the reweighted configuration from $14.01\pm1.94$ to $12.38\pm0.92\%$, so the 3000-epoch operating point stops short of the attainable accuracy on every axis. Already at 3000 epochs the reweighted model's five-seed mean reaches the level of the best submission.

\subsection{Effect of Per-Material Fine-Tuning}
\label{sec:ft}
Any fine-tuning on material~D costs trapezoidal accuracy, and the cost grows with the fine-tuning strength (\Cref{tab:ft}). Starting from the joint model, head-only adaptation of material~D, which updates 697 conditioning and residual parameters with a frozen backbone, already concedes trapezoidal accuracy, $17.1\%$ at seed~0 against $15.6\%$ untouched. Full-parameter adaptation degrades it sharply to $25.2\%$, and a strong head-mode schedule inside the mixed configuration collapses individual seeds to $58.2\%$. The mechanism is structural, and learning-rate tuning does not remove it. Material~D's adaptation set is the trapezoid-free training set itself, so every gradient step on it pulls the FiLM and residual parameters back toward the biased distribution and overwrites the borrowed trapezoidal knowledge, a catastrophic forgetting \cite{mccloskey1989} whose extent grows with the fine-tuning strength. The mixed configuration that fine-tunes only the well-supported materials and leaves material~D untouched, MIST+FT in \Cref{tab:main}, recovers a marginally better aggregate of $7.24$ against $7.32\%$ at the cost of $1.6$ percentage points on material~D and doubled variance. The proposed configuration therefore deploys material~D directly from the joint model.

\begin{table}[!tb]
\caption{Per-Material Fine-Tuning After Joint Pretraining (Seed~0 Unless Noted)}
\label{tab:ft}
\centering
\small
\setlength{\tabcolsep}{3pt}
\renewcommand{\arraystretch}{1.2}
\begin{tabular}{@{}p{3.2cm} c c c@{}}
\toprule
Configuration & Agg.\ $p95$ & D $p95$ & D trap.\ (\%) \\
\midrule
No FT (proposed)$^{\dag}$ & $7.32\pm0.19$ & $12.38\pm0.92$ & $15.16\pm1.69$ \\
FT-head on D$^{\ast}$ & 8.02 & 14.71 & 17.1 \\
FT-full on D & 7.98 & 16.03 & 25.2 \\
Strong FT mix$^{\dag}$ & $7.49\pm0.61$ & $15.23\pm2.32$ & $31.7\pm16.8$ \\
MIST+FT, D untouched$^{\dag}$ & $7.24\pm0.50$ & $14.01\pm1.93$ & $18.08\pm4.18$ \\
\bottomrule
\end{tabular}
\par\vspace{2pt}
{\footnotesize\raggedleft $^{\dag}$Five-seed mean$\pm$std; the remaining rows are seed~0. $^{\ast}$Only the 697 parameters of the head are updated.\par}
\end{table}

\subsection{Zero-Shot Transfer and Data Scaling}
\label{sec:generalization}
Two questions probe the mechanism directly. \emph{How much of the gain is borrowed?} A zero-shot variant removes all eight material-D training trapezoids while retaining its sinusoidal, triangular, and square-like samples. The retention is essential, since those samples calibrate the material's loss scale in \cref{eq:physhead} and its FiLM parameters in \cref{eq:film}. A naive variant that removes material~D entirely fails catastrophically, with $p95$ in the hundreds to thousands of percent depending on the substituted identity, because scale calibration and trapezoidal geometry are then confounded. With the scale anchored and not a single material-D trapezoid seen, the model attains a material-D $p95$ of $15.49\pm1.25\%$ and a trapezoidal $p95$ of $19.8\%$ (\Cref{tab:gen}), on par with MagLearn's sequential-transfer result of $15.9\%$ obtained \emph{with} the eight trapezoids. The accuracy on the material-D test trapezoids therefore rests on measured support borrowed through identity conditioning. The gap to the few-shot reference of $14.01\pm1.94\%$, trained under the same 3000-epoch schedule with the eight trapezoids present, then bounds the total value of material~D's own eight trapezoids at about $1.5$ percentage points: most of the improvement comes from the borrowed support, and only a small share from material~D's own trapezoids.

\emph{Does more borrowed data keep helping?} Extending the pool with ten legacy MagNet materials \cite{magnet2023} adds 23128 trapezoids and widens the identity one-hot to 15. It improves the same-epoch comparison on every axis and reaches at 1500 epochs the level that the five-material pool requires 10000 epochs to attain: D $14.17\pm0.54\rightarrow12.57\pm1.88\%$, trapezoidal $18.8\rightarrow16.6\%$, and aggregate $9.51\rightarrow8.36\%$. The gain saturates because the sibling materials already cover $99.3\%$ of the target support (\Cref{sec:premise}): scaling follows support coverage, not raw data volume, and its returns diminish once the target support is covered.

\begin{table}[!tb]
\caption{Zero-Shot Transfer and Data Scaling on Material D}
\label{tab:gen}
\centering
\footnotesize
\setlength{\tabcolsep}{2pt}
\renewcommand{\arraystretch}{1.2}
\begin{tabular}{@{}p{3.2cm} c c c@{}}
\toprule
Configuration & D $p95$ & D trap.\ & Agg.\ \\
\midrule
Zero-shot$^{\ast}$ & $15.49\pm1.25$ & 19.8 & 8.78 \\
Few-shot reference$^{\dag}$ & $14.01\pm1.94$ & $18.1\pm4.3$ & $8.60\pm0.46$ \\
A--E pool only$^{\ddag}$ & $14.17\pm0.54$ & 18.8 & 9.51 \\
$+$10 legacy materials$^{\ddag}$ & $12.57\pm1.88$ & 16.6 & 8.36 \\
\textbf{Proposed}$^{\S}$ & $\mathbf{12.38\pm0.92}$ & $\mathbf{15.16\pm1.69}$ & $\mathbf{7.32\pm0.19}$ \\
\bottomrule
\end{tabular}
\par\vspace{2pt}
{\footnotesize\raggedleft $^{\ast}$No material-D trapezoid in training, four seeds, 3000 epochs. $^{\dag}$3000 epochs, $K{=}5$. $^{\ddag}$1500 epochs, $K{=}5$. $^{\S}$10000 epochs, $K{=}5$, i.e., the configuration of \Cref{tab:main}.\par}
\end{table}

\subsection{Discussion}
\emph{Why borrowed support succeeds where sequential transfer stalled.} Every remedy confined to material~D's own samples leaves the trapezoidal axis unmoved or worse (\Cref{sec:failures}), whereas measured support borrowed from the sibling materials roughly halves it, even in the zero-shot setting (\Cref{sec:generalization}). External measured data are therefore the contribution under test: because no manipulation of material~D's own 580 samples restores the missing geometry (\Cref{sec:failures}), the comparison against the single-material baseline measures what borrowed support adds, and every table marks the data regime. The sequential-transfer entries of the challenge consumed the same kind of external data and still left material~D at $15.9\%$ or worse. They fold external knowledge into initial weights and then adapt on the target material's own training set, whereas MIST keeps all materials in one objective, separates them through identity conditioning, and never adapts the scarce material on the set that lacks the target geometry (\Cref{sec:ft}). Wherever sibling materials are characterized on the same excitation hardware, the common case in magnetics, the joint regime is available to any model family at no measurement cost.

\emph{Transferable principles.} Three rules follow. When share matching does not help, the difficulty is missing information and the imbalance toolbox does not apply. Labels should come only from measurement or from an exact symmetry, because an invented label turns prior bias into shared-weight pollution. Missing support is then borrowed from external measured data through identity conditioning, and the scarce material is not fine-tuned on its own biased set.

\emph{Limitations and future work.} The evidence covers the material-D instance, where the diagnosis, the mechanism, and the result are each independently controlled. Whether identity-conditioned transfer repairs \emph{any} designed waveform hole remains to be established on further materials with shifts of comparable severity. All materials studied are ferrites; transfer across material classes with different loss mechanisms, such as nanocrystalline or powder cores, is untested. Eliminating the mild concession of the well-supported materials (\Cref{sec:main}) and pushing material~D into single digits remain open.

\section{Conclusion}
The open waveform challenge of the MagNet benchmark is a designed instance of the characterization--deployment mismatch, and this paper has diagnosed it experimentally as a missing-information problem under covariate shift. Four single-material remedies fail, and share matching makes the trapezoidal error worse, so the missing loop geometry has to come from measured data of sibling materials. MIST supplies it with a single 2784-parameter predictor trained jointly on five materials: material identity enters through FiLM conditioning, the scarce material's true-label loss is reweighted, and material~D is never fine-tuned. The material-D $p95$ falls from $20.39\pm2.03\%$ to $12.38\pm0.92\%$, below the best challenge submission at one-sixth of its parameters, and the five-material aggregate improves as well. A capacity-matched ablation isolates the mechanism: without identity information the same data collapse by an order of magnitude, so conditioning, not capacity or volume, is the interface through which external support transfers. Leaving the scarce material without fine-tuning keeps that borrowed support intact. More generally, wherever characterization data and deployment waveforms diverge, we argue that the scarce material should be diagnosed before it is remedied, and that its missing support should be transferred from sibling materials through identity conditioning rather than re-fitted on its own biased set.

\bibliographystyle{IEEEtran}
\bibliography{reference}

\end{document}